\documentclass[review]{elsarticle}

\usepackage{lineno,hyperref}
\usepackage{amsmath}
\usepackage{times}
\usepackage{epsfig}
\usepackage{graphicx}
\usepackage{amsmath}
\usepackage{amssymb}

\usepackage{caption}
\usepackage{algorithm}
\usepackage{algorithmic}
\usepackage{float}
\usepackage{amsmath}
\usepackage{mathrsfs}

\journal{Journal of \LaTeX\ Templates}

\begin{document}

\begin{frontmatter}

\title{Class label autoencoder for zero-shot learning}

\author[mymainaddress]{Guangfeng Lin\corref{mycorrespondingauthor}}
\cortext[mycorrespondingauthor]{Corresponding author}
\ead{lgf78103@xaut.edu.cn}


\author[mymainaddress]{Caixia Fan}
\author[mymainaddress]{Wanjun Chen}
\author[mymainaddress]{Yajun Chen}
\author[mymainaddress]{Fan Zhao}

\address[mymainaddress]{Information Science Department, Xi’an University of Technology,\\
 5 South Jinhua Road, Xi'an, Shaanxi Province 710048, PR China}


\begin{abstract}
Existing zero-shot learning (ZSL) methods usually learn a projection function between a feature space and a semantic embedding space(text or attribute space) in the training seen classes or testing unseen classes. However, the projection function cannot be used between the feature space and  multi-semantic embedding spaces, which have the diversity characteristic for describing the different semantic information of the same class. To deal with this issue, we present a novel method to ZSL based on learning class label autoencoder (CLA). CLA can not only build a uniform framework for adapting to multi-semantic embedding spaces, but also construct the encoder-decoder mechanism for constraining the bidirectional projection between the feature space and the class label space. Moreover, CLA can jointly consider the relationship of feature classes and the relevance of the semantic classes for improving zero-shot classification. The CLA solution can provide both unseen class labels and the relation of the different classes representation(feature or semantic information) that can encode the intrinsic structure of classes. Extensive experiments demonstrate the CLA outperforms state-of-art methods on four benchmark datasets, which are AwA, CUB, Dogs and ImNet-2.
\end{abstract}


\end{frontmatter}

\section{Introduction}
The large-scalely visual recognition problem can be solved possible by the support of large-scale datasets (for example, ImageNet \cite{Russakovsky2015ImageNet}) and the advances of deep learning methods \cite{Krizhevsky2012ImageNet} \cite{Sermanet2013OverFeat} \cite{Simonyan2014Very} \cite{Szegedy2015Going}. However, the visual recognition is still challenging "in the wild" because of the rare samples object classes and fine-grained object categories. For example, we cannot always collect all classes, which include the recognized classes, for learning the related model. Moreover, we can hardly utilize the model learned based on coarse classes to cognise fine-grained classes in traditionally visual recognition methods. To deal with the visual recognition problem in these situations, the main idea of ZSL is to exploit the transfer model between the feature space and the semantic space in seen classes, form which labeled samples can be used, for classifying unseen classes, from which samples can not be collected. In other words, in ZSL, training and testing class sets are disjoint which can be handled by modeling the transfer relationship based on the interactive relevance between feature classes and semantic classes. For instance, 'horse' belongs to unseen classes, while 'zebra' is a seen class. These classes include the same (e.g. 'horse' and 'zebra' are both 'has tail' attribute) or different semantic information (e.g. zebra has 'is striped' attribute, but horse has 'is solid color' attribute). In ZSL, the knowledge transfer model can be learned between the visual feature of 'zebra' and the semantic information of 'has tail' in training sets, and then, in classification, the visual feature of 'horse' can be projected into the semantic or label space by the transfer model for recognizing 'horse' in testing sets.

In ZSL, the transfer model can be learned between the visual feature and the semantic information in training seen class sets, so that it suffers from project domain shift problem \cite{Fu2015Transductive} \cite{Kodirov2017} because of the disjoint between testing unseen class sets and training seen class sets. To alleviate the effect of the project domain shift problem, we face to two challenges\cite{Changpinyo2016}. One is how to use the semantic information for modeling the knowledge transfer relationship, and other is how to construct the projection function between visual feature and multi-semantic information for the optimally discriminative characteristic on unseen classes.

To address the first challenge, we usually make a assumption, in which seen and unseen classes are correlative in the semantic embedding space (e.g. attribute \cite{farhadi2009describing} \cite{lampert2009learning} \cite{parikh2011relative} or text space \cite{Frome2013DeViSE} \cite{mikolov2013efficient} \cite{Socher2013Zero}). In the semantic embedding space, all kinds of class names are embedded as vectors, which are class prototype \cite{Fu2014Transductive}. Some ZSL methods \cite{Fu2014Transductive} \cite{jayaraman2014zero} \cite{Lampert2014} \cite{li2015semi} \cite{li2014attributes} \cite{norouzi2013zero} \cite{romera2015embarrassingly} try to find the relation of the different space (feature, semantic embedding and class label space) for modeling the knowledge transfer, and others attempt to transform the semantic embedding into new representation for constraining the transfer relationship between seen and unseen classes by Canonical Correlation Analysis (CCA)\cite{Fu2015Transductive} or Sparse Coding (SC) \cite{yu2017transductive} \cite{zhang2015zero}.

To handle the second challenge, attribute-based classification \cite{Lampert2014} as the classical method can construct the probability model to predict the visual attributes of unseen classes. Recent methods tend to build the linear \cite{Frome2013DeViSE} \cite{7298911} \cite{Akata2016Label} \cite{Kodirov2017}, nonlinear \cite{Socher2013Zero} \cite{7780384}, or hybrid \cite{norouzi2013zero} \cite{zhang2015zero} projection function among the different spaces (feature, semantic embedding, and class label space). Furthermore, two tendencies show the promising results. One is that the structure of semantic classes \cite{Changpinyo2016} or structure propagation \cite{Lin2018structure} are considered for enhancing the transfer model based on the above projection function. The other is a autoencoder mechanism is utilized to constrain the bidirection projection relation between feature and semantic information for improving the compatibility of the transfer model\cite{Kodirov2017}. However, this autoencoder mechanism neglects the model construction between the feature space and multi-semantic space. Moreover, it is difficult to approximate the intrinsic structure of the unseen classes because of the statically linear model learned on seen classes.

\begin{figure*}[ht]
  \begin{center}
\includegraphics[width=1\linewidth]{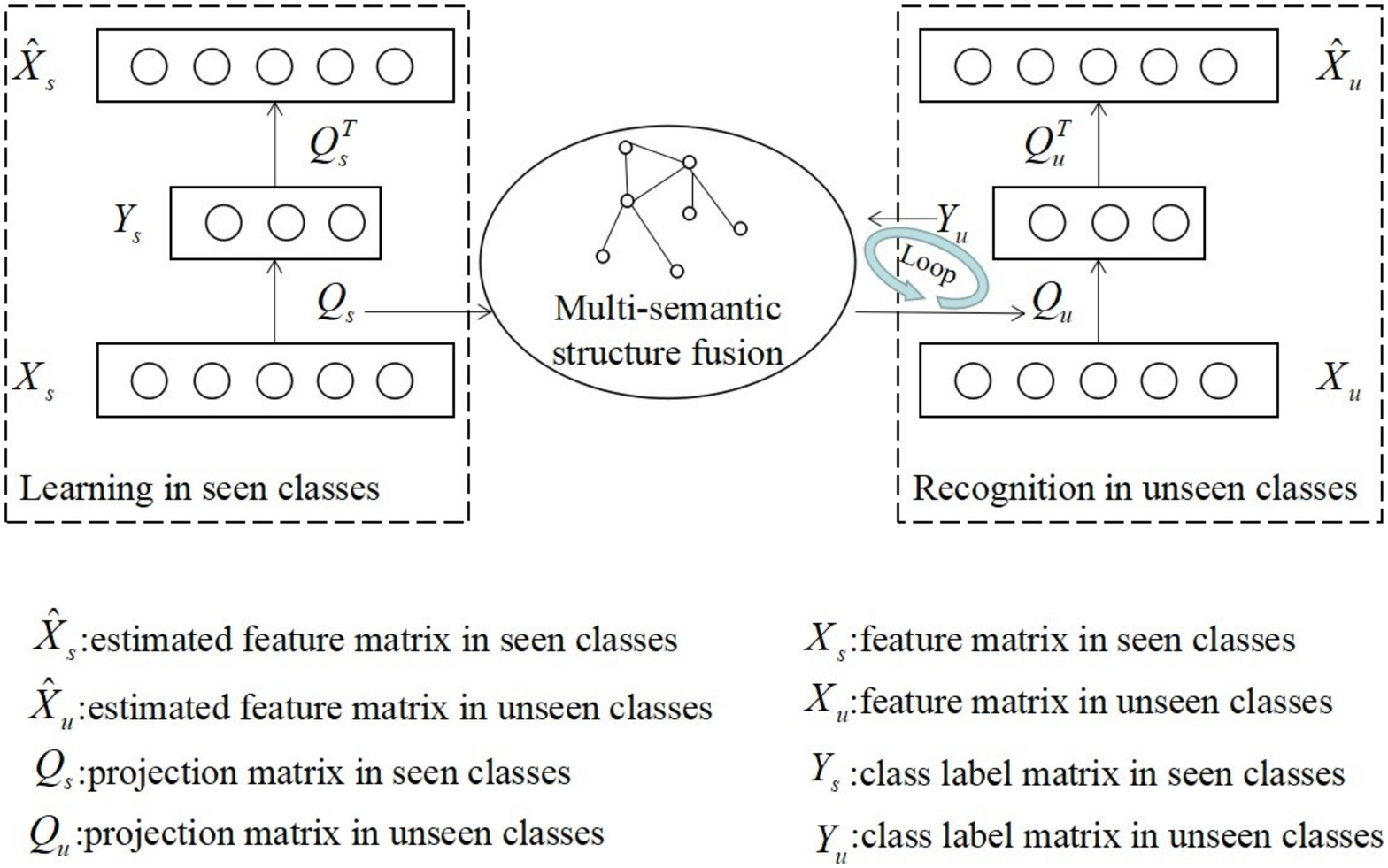}
\end{center}
\vspace{-0.2in}
 \caption{The illustration of class label autoencoder for zero-shot learning.}
  \label{figmot}
 \end{figure*}

From above mentions, our motivation is inspired by the autoencoder mechanism \cite{Kodirov2017}, structure fusion \cite{Lin2017275} \cite{Lin20161} \cite{Lin2014146} \cite{7268821} \cite{7301305} \cite{Lin20131286} and structure propagation \cite{Lin2017Dynamic} \cite{Lin2018structure} for jointly addressing two challenges. The different point of CLA try to model the projection function between the feature and the class label space by autoencoder mechanism for dealing with the transfer relationship among the feature and multi-semantic embedding spaces, while literature \cite{Kodirov2017} only involves the model construction between the feature and the single semantic space, and literature \cite{Lin2018structure} can not consider the bidirectional constrain between the feature and the class label space with the consideration of multi-semantic information for building the transfer model. Therefore, we expect that CLA not only can construct a uniform framework for adapting to multi-semantic embedding spaces, but also reform the encoder-decoder mechanism for constraining the bidirectional projection between the feature space and the class label space. Figure \ref{figmot} illustrates the idea of CLA conceptually.

Our main contributions have two points. One is a novel idea proposed to construct projection function between the feature and the class label space based on autoencoder mechanism for considering the multi-semantic information in ZSL. Other is a feasible method presented to improve unseen classes recognition by the evolution relationship mining of the different manifold structure(the feature distribution structure of seen classes,the feature distribution structure of unseen classes, the semantic distribution structure between seen and unseen classes). In the experiment, we evaluate CLA on four benchmark datasets for ZSL. The experimental results are promising for ameliorating the recognition performance of unseen object classes.

\section{Related Works}
In ZSL, the semantic information(For example, attributes \cite{lampert2009learning} come from the manual annotation \cite{Akata2013Label}and text information \cite{Socher2013Zero} derive from language by machine learning \cite{Yu2013Designing} or data mining \cite{Elhoseiny2013Write}) can describe the characteristic of each class corresponding to the class label. Recent methods can draw support from the semantic information to recognize the visual unseen class by intermediate attribute classifiers learning \cite{Rohrbach2010What} \cite{Rohrbach2011Evaluating} \cite{Lampert2014}, seen class proportions combination \cite{zhang2015zero} \cite{7781018} \cite{norouzi2013zero} \cite{Changpinyo2016}, and compatibility learning \cite{Akata2016Label} \cite{7298911} \cite{Frome2013DeViSE} \cite{Socher2013Zero} \cite{romera2015embarrassingly} \cite{7780384}. For directly classifying unseen classes, the current focus is how to learning the projection or compatibility function between the feature space and the class label space by the assistance of the semantic information to suppress the project domain shift problem. We attempt to enhance the projection or compatibility function learning by structure fusion of multi-semantic classes and autoencoder mechanism, which are the related methods of CLA.

To best of our knowledge, structure fusion is firstly proposed for multi-feature fusion of shape classification \cite{Lin20131286} in our research. Structure is defined as the graph structure among features of data samples. We can capture the linear relation of multi-feature structure fusion base on manifold leaning method \cite{Lin20131286} \cite{Lin2014146} or probability model \cite{7268821} for improving the performance of image classification, shape analysis and human action recognition. Moreover, in the further works, we construct the non-linear relation of heterogeneous structures fusion for remarkably ameliorating image classification \cite{7301305} \cite{Lin20161} and feature encoding \cite{Lin2017275}. In recent works, we find the interesting things that are dynamic structure fusion to refining the relation of objects for semi-supervised multi-modality classification \cite{Lin2017Dynamic} and structure propagation to update the relevance of multi-semantic classes by the iteration computation for ZSL \cite{Lin2018structure}. From above works, it shows that the relationship between structures is very important for the discriminative learning of object classification. Therefore, we expect to deal with structure fusion of multi-semantic classes for ZSL by the more suitable way.

Autoencoder is a bidirectional mechanism for encoding and decoding in many works \cite{Baldi1989Hornik} \cite{Rifai2011Contractive} \cite{Xie2015Unsupervised} \cite{Chen2012Marginalized} \cite{Badrinarayanan2017SegNet}\cite{Yan2015Attribute2Image} \cite{Reed2016Generative} \cite{Kodirov2017}. In term of semantic projection, autoencoder methods are roughly divided into two categories which are non-semantic and semantic encoder-decoder methods. Non-semantic encoder-decoder methods usually learn the intrinsic structure of data for visualization/clustering \cite{Xie2015Unsupervised} or classification \cite{Chen2012Marginalized}, while semantic encoder-decoder methods generally share the latent embedding space between the encoder and the decoder by semantic regularization \cite{Yan2015Attribute2Image} \cite{Reed2016Generative} or learn end-to-end deep model for ZSL by reconstructing the loss between the convolutional and deconvolutional neural network \cite{Kodirov2017}.

For considering the bidirectional constrains, we expect to build the project model between visual features and class labels for ZSL by autoencoder idea. Simultaneously, we want to use structure fusion idea to process multi-semantic information by the model for improve the performance of ZSL. To this end, we propose CLA for uniforming two ideas in ZSL.
\section{Class Label Autoencoder}
\subsection{Linear autoencoder}
linear autoencoder can construct a simple model, which only includes one hidden layer linked between the encoder and decoder. By this mechanism, the input data can be encoded by projecting into the hidden layer, and then can be decoded by reconstructing the original data space \cite{Kodirov2017}. Therefore, linear autoencoder can attain the better coding quality with minimizing the error between the original and reconstructed data. We extend the autoencoder mechanism into class label space and expect to directly encode the visual feature to the class label with the semantic information. Given, an input data is a visual feature matrix $X\in \mathbb{R}^{d\times N}$ ($N$ feature vectors of $d$ dimensions), and can be projected into a $k$-dimension class label space by a transformation matrix $Q \in \mathbb{R}^{k\times d}$. The class label representation is $Y \in \mathbb{R}^{k \times N}$. According to autoencoder mechanism, the class label representation can be mapped back to become $\hat{X} \in \mathbb{R}^{d \times N}$ by a transformation matrix $Q^{*} \in \mathbb{R}^{d\times k}$. The tied weights can be considered to further simplify the autoencoder model by $Q^{*}=Q^{T}$ \cite{Ranzato2007Sparse}. We expect to minimize the reconstruct error between $\hat{X}$ and $X$. To this end, the following objective can be built.
 \begin{align}
\label{objective1}
\begin{aligned}
&Q=\arg \min_{Q}\|X-Q^{T}QX\|^{2}_{F},&s.t.~~~~QX=Y
 \end{aligned}
\end{align}
We can equivalently reformulate (\ref{objective1}) as unconstrained optimization problem as following.
 \begin{align}
\label{objective11}
\begin{aligned}
&Q=\arg \min_{Q}\|X-Q^{T}Y\|^{2}_{F}+\lambda\|QX-Y\|^{2}_{F}
 \end{aligned}
\end{align}
here,$\lambda$ is a tradeoff parameter for balancing the encoder and the decoder.
\subsection{ZSL Model}
In ZSL, visual features include two sets. One set can be represented as feature matrix $X_{s}\in \mathbb{R}^{d\times N_{s}}$ ($N_{s}$ feature vectors of $d$ dimensions) with class label matrix $Y_{s} \in \mathbb{R}^{k_{s}\times N_{s}}$ ($N_{s}$ label vectors of $k_{s}$ dimensions) in seen classes, and another can be described as feature matrix $X_{u}\in \mathbb{R}^{d\times N_{u}}$ ($N_{u}$ feature vectors of $d$ dimensions) without class label matrix $Y_{u}\in \mathbb{R}^{k_{u}\times N_{u}}$ ($N_{u}$ label vectors of $k_{u}$ dimensions) in unseen classes. Semantic feature set can be defined as $S=\{S_{s},S_{u}\}$, in which $S_{s}\in \mathbb{R}^{d_{s}\times k_{s}}$ or $S_{u}\in \mathbb{R}^{d_{s}\times k_{u}}$ respectively is feature matrix in seen or unseen classes. We expect to learn a transformation matrix $Q_{s} \in \mathbb{R}^{k_{s}\times d}$ in seen classes for transferring knowledge to a transformation matrix $Q_{u}\in \mathbb{R}^{k_{u}\times d}$ in unseen classes with consideration of multi-semantic information. Therefore, we want to find the relationship between $Q_{s}$ and $Q_{u}$. In term of this relationship, the efficient information can be transferred from seen classes to unseen classes. To this end, we respectively define the following formula.
\begin{align}
\label{objective2}
\begin{aligned}
Q_{s}=W_{s}^{T}A_{s},
  \end{aligned}
\end{align}
here, $W_{s}\in\mathbb{R}^{k_{s}\times k_{s}} $ ($k_{s}$ is the number of seen classes)is the similarity matrix of seen classes, and is the structure representation of seen classes. $A_{s} \in \mathbb{R}^{k_{s}\times d}$ is a projection matrix for encoding seen classes. By formula (\ref{objective2}), we can decompose $Q_{s}$  into two parts, in which $W_{s}$ can describe the intrinsically discriminative characteristic of seen classes and $A_{s}$ can extract the common information in seen classes. In unseen classes, we define the similar formula as following.
\begin{align}
\label{objective3}
\begin{aligned}
Q_{u}=W_{u}^{T}A_{u},
  \end{aligned}
\end{align}
here, $W_{u}\in\mathbb{R}^{k_{u}\times k_{u}} $ ($k_{u}$ is the number of unseen classes)is the similarity matrix of unseen classes, and is the structure representation of unseen classes. $A_{u}\in \mathbb{R}^{k_{u}\times d}$ is a projection matrix for encoding unseen classes. By  formula (\ref{objective3}), we can decompose $Q_{u}$  into two parts, in which $W_{u}$ can depict the intrinsically discriminative characteristic of unseen classes and $A_{u}$ can extract the common information in unseen classes. For describing the transfer relationship of the common information between seen and unseen classes, we define the following formula.
\begin{align}
\label{objective4}
\begin{aligned}
A_{u}=W_{su}^{T}A_{s},
  \end{aligned}
\end{align}
here, $A_{u}$ can not directly  be obtained by autoencoder mechanism. Therefore, we compute $A_{u}$ by $A_{s}$ and $W_{su}\in\mathbb{R}^{k_{s}\times k_{u}}$,which is the similarity matrix and transfers the common information between seen and unseen classes. In formula (\ref{objective2}),(\ref{objective3}) and (\ref{objective4}), the similarity matrix can be computed as following.
 \begin{align}
\label{weight}
\begin{aligned}
w_{ij}=\frac{\exp (-d(z_{i},z_{j}))}{\sum_{i=1,j=1}^{n_{i},n_{j}}\exp (-d(z_{i},z_{j}))},
 \end{aligned}
\end{align}
 \begin{align}
\label{distance}
\begin{aligned}
d(z_{i},z_{j})=(z_{i}-z_{j})^{T}\Sigma_{z}^{-1}(z_{i}-z_{j}),
 \end{aligned}
\end{align}
here, when $z_{i}$ and $z_{j}$ respectively are visual class feature or semantic class representation in seen classes, $w_{ij}$ is the element of $W_{s}$; while $z_{i}$ and $z_{j}$ respectively are visual class feature or semantic class representation in unseen classes, $w_{ij}$ is the element of $W_{u}$; if $z_{i}$ and $z_{j}$ respectively are visual class feature or semantic class representation in seen or unseen classes, $w_{ij}$ is the element of $W_{su}$.

According to the above definitions, we can equivalently reformulate formula (\ref{objective11}) in seen classes as following.
 \begin{align}
\label{objective5}
\begin{aligned}
&A_{s}=\arg \min_{A_{s}}\|X_{s}-A_{s}^{T}W_{s}Y_{s}\|^{2}_{F}+\lambda\|W_{s}^{T}A_{s}X_{s}-Y_{s}\|^{2}_{F}
 \end{aligned}
\end{align}
For optimizing formula (\ref{objective5}), we can transform it as a well-know Sylvester equation (Appendix A shows the detail). Bartels-Stewart algorithm \cite{Bartels1972Solution} can be efficiently solve this problem by running Matlab function "sylvester". When we obtain $A_{s}$, $A_{u}$ can be computed by formula (\ref{objective2}) and (\ref{objective4}).  In term of formula (\ref{objective3}), $Y_{u}$ can be calculated by the following formula.
 \begin{align}
\label{objective6}
\begin{aligned}
&Y_{u}=Q_{u}X_{u}=W_{u}^{T}W_{su}^{T}A_{s}X_{u}
 \end{aligned}
\end{align}
We can determinate a estimated label $\hat{y}$ to a $x_{u} \in X_{u}$ by the following formula.
 \begin{align}
\label{objective7}
\begin{aligned}
&\hat{y}=\arg \max_{1\leq c \leq k_{u}}W_{u}^{T}W_{su}^{T}A_{s}x_{u}
 \end{aligned}
\end{align}
\subsection{Multi-semantic structure evolution fusion}
In formula (\ref{objective5}),(\ref{objective6}) and (\ref{objective7}), the similarity matrix ($W_{s}$, $W_{u}$ or $W_{su}$) can be computed by formula (\ref{weight}). In ZSL, semantic information can usually describe all class prototype. Therefore, the similarity matrix often is the structure of the semantic information. If we can obtain multi-semantic information (for example, attribute\cite{lampert2009learning} or word vector\cite{Socher2013Zero}), the structure of the semantic information can be modeled by the linear relationship of multi-similarity matrix in the following formula.
\begin{align}
\label{objective8}
\begin{aligned}
&W_{s}=\sum^{M}_{i=1}\alpha_{i}W_{(s,i)}\\
&W_{u}=\sum^{M}_{i=1}\alpha_{i}W_{(u,i)}\\
&W_{su}=\sum^{M}_{i=1}\alpha_{i}W_{(su,i)}\\
&s.t.~~~~\sum^{M}_{i=1}\alpha_{i}=1
 \end{aligned}
\end{align}
here, $M$ is the number of multi-semantic information. $\alpha_{i}$ is the linear coefficient of the similarity matrix, which is $W_{(s,i)}$ (the structure representation of $i$th semantic information in seen classes), $W_{(u,i)}$(the structure representation of $i$th semantic information in unseen classes) or $W_{(us,i)}$(the structure representation of $i$th semantic information between seen and unseen classes).

Except multi-semantic information, visual feature also can help to construct the structure representation of object classes(A kind of method is simple for describing visual class by averaging all visual features with the same label). However, we only know the label of unseen classes by formula (\ref{objective7}). In other words, we expect to calculate the structure representation of visual feature class by the estimated label. Therefore,the above similarity matrix come from not only semantic information but also visual feature. We reformulate formula (\ref{objective8}) as follows.
\begin{align}
\label{objective9}
\begin{aligned}
&W_{s}=\sum^{M}_{i=1}\beta_{i}W_{(s,i)}+\beta_{M+1}W_{(s,I)}\\
&W_{u}=\sum^{M}_{i=1}\gamma_{i}W_{(u,i)}+\gamma_{M+1}W_{(u,I)}\\
&W_{su}=\sum^{M}_{i=1}\gamma_{i}W_{(su,i)}+\gamma_{M+1}W_{(su,I)}\\
&s.t.~~~~\sum^{M+1}_{i=1}\beta_{i}=1,\sum^{M+1}_{i=1}\gamma_{i}=1
 \end{aligned}
\end{align}
here,$W_{(s,I)}$, $W_{(u,I)}$ and $W_{(su,I)}$ are respectively the structure representation of visual feature class in seen classes, unseen classes and between seen and unseen classes. $W_{(s,I)}$ is fix because of the label in seen classes, while $W_{(u,I)}$ and $W_{(su,I)}$ are dynamic with the estimated label in unseen classes. Therefore, we respectively use $\beta=[\beta_{1}~~\beta_{2}~~ ...~~ \beta_{M+1}]^{T}$ and $\gamma=[\gamma_{1}~~\gamma_{2}~~ ...~~ \gamma_{M+1}]^{T}$ for weight the different structure representation. We can reformulate formula (\ref{objective5}) in term of formula (\ref{objective9}) as follows.
\begin{align}
\label{objective10}
\begin{aligned}
&(A_{s}, \beta)=\arg \min_{A_{s}}\|X_{s}-A_{s}^{T}W_{s}Y_{s}\|^{2}_{F}+\lambda\|W_{s}^{T}A_{s}X_{s}-Y_{s}\|^{2}_{F}
 \end{aligned}
\end{align}
For optimizing formula (\ref{objective10}), we fix $\beta$ to optimize $A_{s}$ by transforming it as a well-know Sylvester equation (Appendix A shows the detail), and then fix $A_{s}$ to solve $\beta$ by linear programming. We can obtain the initial estimated label by formula (\ref{objective6})and (\ref{objective7}). The initial estimated label could not be accurate, so a evolution process can be presented to refine the performance of the model by the iteration computation. Therefore, we can update $W_{(u,I)}$ and $W_{(su,I)}$ by the estimated label. For updating $\gamma$, we construct formula (\ref{objective12}) to constrain the positive propagation of label.
 \begin{align}
\label{objective12}
\begin{aligned}
\gamma=\arg \min_{\gamma}(&\|\sum^{M}_{i=1}\gamma_{i} W_{(u,i)}- \gamma_{M+1}W_{(u,I)}\|^{2}_{F}+\\
&\delta\|\sum^{M}_{i=1}\gamma_{i}W_{(su,i)}-\gamma_{M+1} W_{(su,I)}\|^{2}_{F})\\
=\arg \min_{\gamma}(&\|\gamma^{T}P_{u}\|^{2}_{F}+\delta\|\gamma^{T}P_{su}\|^{2}_{F})\\
&s.t.~~~~\sum^{M+1}_{i=1}\gamma_{i}=1
 \end{aligned}
\end{align}
here, The element of $W_{(u,i)}$ and $W_{(u,I)}$ forms the column of $P_{u}\in \mathbb{R}^{(M+1)\times (k_{u}\times k_{u})}$, while the element of $W_{(su,i)}$ and $W_{(su,I)}$ makes the column of $P{su}\in \mathbb{R}^{(M+1)\times (k_{s}\times k_{u})}$.
Formula (\ref{objective12}) can be transformed as a linear programming for solving $\gamma$. The evolution process can be implemented in formula (\ref{objective9}),(\ref{objective6}),(\ref{objective7}) and (\ref{objective12}) by the iteration computation.

To describe the detail of CLA, we demonstrate the pseudo code of the proposed CLA algorithm in Algorithm \ref{algCLA}, which includes three steps. The first step (line 1 and line 2) initializes the structure representation of unseen classes and the structure relationship. The second step (line 3 and line 4) computes the fusion structure for computing a projection matrix of seen classes and updating the structure relationship. The third step (from line 6 to line 11) is a evolution process for refine the classification performance of unseen classes by iteration computation. In addition, the evolution process can also fuse the recognition result of other ZSL method for further improving the classification performance of unseen classes.
\begin{algorithm}[ht]
  \caption{The pseudo code of the CLA algorithm}
 \begin{algorithmic}[1]
 \label{algCLA}
\renewcommand{\algorithmicrequire}{\textbf{Input:}}
\renewcommand{\algorithmicensure}{\textbf{Output:}}
\renewcommand{\algorithmicreturn}{\textbf{Iteration:}}
   \REQUIRE $X_{s}$,$X_{u}$,$Y_{s}$,$S_{s}$ and $S{u}$
   \ENSURE $\hat{y}_{P}$ ($P$ is the total iteration number )
   \STATE Setting the similarity matrix $W_{(u,I)}$ and $W_{(su,I)}$ to zero matrix
   \STATE Setting the initial value of every element is $1/(M+1)$ in $\gamma$ and $\beta$
   \STATE Computes the similarity matrix $W_{s}$,$W_{u}$ and $W_{su}$ by (\ref{objective9}) and (\ref{weight})
   \STATE Solving $A_{s}$ and updating $\beta$ by alternately optimizing (\ref{objective10})
   \STATE Estimating the label of unseen classes according to (\ref{objective6}) and (\ref{objective7})
   \FOR {$1<t<P$}
   \STATE Updating $W_{(u,I)}$ and $W_{(su,I)}$ by (\ref{weight}) and the estimated label.
   \STATE Updating $\gamma$ by optimizing (\ref{objective12})
   \STATE Updating the similarity matrix $W_{u}$ and $W_{su}$ by (\ref{objective9})
   \STATE Estimating the label $\hat{y}_{t}$ of unseen classes according to (\ref{objective6}) and (\ref{objective7})
   \ENDFOR
  \end{algorithmic}
\end{algorithm}
\section{Experiment}
\subsection{Datasets}
We evaluate the proposed method CLA in four challenging datasets, which include Animals with Attributes (AwA)\cite{Lampert2014}, CUB-200-2011 Birds (CUB)\cite{Wah2011The}, Stanford Dogs (Dogs)\cite{Deng2013Fine}, and ILSVRC2012/ILSVRC2010 (ImNet-2)\cite{Russakovsky2015ImageNet}. In ImNet-2, the same configuration as in \cite{Kodirov2017} is the 1000 classes of ILSVRC2012 for seen classes and the 360 classes of ILSVRC2010 for unseen classes. These datasets can be categorized into fine-grained recognition (CUB and Dogs) or non-fine-grained recognition (AwA and ImNet-2) for ZSL. Tab.\ref{table1} shows the statistics and the extracted features (the detail of image and semantic feature in section \ref{feature}) for these datasets.
\begin{table*}[!ht]
\small
\renewcommand{\arraystretch}{1.0}
\caption{Datasets statistics and the extracted feature in experiments.}
\label{table1}
\begin{center}
\newcommand{\tabincell}[2]{\begin{tabular}{@{}#1@{}}#2\end{tabular}}
\begin{tabular}{lp{1.2cm}p{1.6cm}p{1.5cm}p{2cm}p{2cm}p{0.5cm}}
\hline
\bfseries Datasets & \bfseries \tabincell{l}{Number of \\seen classes} & \bfseries \tabincell{l}{Number of \\unseen classes} & \bfseries \tabincell{l}{Total number \\of images} & \bfseries \tabincell{l}{Semantic feature\\/dimension} &\bfseries \tabincell{l}{Image feature\\/dimension}\\
\hline \hline
AwA  & $40$ &$10$& $30473$ & \tabincell{l}{att/85,\\w2v/400,\\glo/400,\\hie/about 200.} & \tabincell{l}{Deep feature based\\ on GoogleNet\cite{7298911}\\/1024}\\
\hline
CUB  & $150$ &$50$& $11786$ & \tabincell{l}{att/312,\\w2v/400,\\glo/400,\\hie/about 200.} & \tabincell{l}{Deep feature based\\ on GoogleNet\cite{7298911}\\/1024}\\
\hline
Dogs  & $85$ &$28$& $19499$ & \tabincell{l}{N/A,\\w2v/400,\\glo/400,\\hie/about 200.} & \tabincell{l}{Deep feature based\\ on GoogleNet\cite{7298911}\\/1024}\\
\hline
ImNet-2  & $1000$ &$360$& $218000$ & \tabincell{l}{N/A,\\w2v/1000,\\N/A,\\N/A.} & \tabincell{l}{Deep featurebased\\ on GoogleNet\cite{7298911}\\/1024}\\
\hline
\end{tabular}
\end{center}
\end{table*}
\subsection{Image and semantic feature}
\label{feature}
Image and semantic feature description are necessary for modeling ZSL. Because deep feature can capture the discriminative characteristic of objects based on large scale database, we adopt image feature to be the outputs (1024 dimension feature vector) of the pre-trained GoogleNet\cite{7298911} \cite{Szegedy2015Going}, which is end to end paradigm for processing whole image inputs. Semantic feature can be extracted by four methods in the different datasets. The first method obtains the feature vector from attributes (att)\cite{farhadi2009describing} by human annotation and judgment in AwA and CUB. The second method extracts word vectors (w2v) by predicting words of text document on a two-layer neural network \cite{Mikolov2013Distributed} in AwA, CUB, ImNet-2 and Dogs.  The third method attains GloVe (glo) form co-occurrence statistics of words on a large unlabel text corpora \cite{Pennington2014Glove} in AwA, CUB, and Dogs. The forth method gets hierarchical embedding (hie) from vectorial class structure for describing the class hierarchical relationship (for example WordNet \cite{7298911}\cite{Miller2002WordNet}) in AwA, CUB, and Dogs.

\subsection{Classification and validation protocols}
Classification accuracy can be computed by averaging all test classes accuracy in each database. In the learned model, there are three parameters, which are $\lambda$ (the tradeoff parameter and in formula (\ref{objective10}),$P$ (the total iteration number in Algorithm \ref{algCLA}), and $\delta$ (the tradeoff parameter and in formula (\ref{objective12}). The training classes set is alternately segmented as learning set and validation set in according with the proportion between the training classes set and the test classes set. We obtain $\lambda$ corresponding to the best result in $0.001,0.01,0.1, 1, 10, 100, 1000, 10000$ by cross validation. In all experiments, $P$ is set to $50$ and $\delta$ is equal to $0.1$.

\subsection{Comparison with baseline approaches}
In this section, because autoencoder mechanism and structure propagation are basic ideas for constructing CLA, we implement two existing methods with these ideas as baseline approaches, which are  semantic autoencoder (SAE)\cite{Kodirov2017} and structure propagation (SP) \cite{Lin2018structure}. SAE have two configurations, which encode from visual to semantic space(V to S) or from semantic to visual space (S to V). The details of experimental results are illustrated in Tab.\ref{table2}. In the differently semantic space, the classification performance of CLA is obviously superior to that of the baseline methods. In AWA, CUB, Dogs and ImNet-2, the performance of CLA respectively improves $2\%$, $3.7\%$, $3.2\%$, and $0.1\%$ at least. Because the performance improvement is not significant in ImNet-2, we demonstrate the comparison of Top-n (n is a number of set, which includes 1,2,3,4 and 5.) accuracy between CLA and the baseline methods on unseen classes in Tab.\ref{table3}. CLA can still obtain the best performance in the contrast methods.

\begin{table}[!ht]
\small
\renewcommand{\arraystretch}{1.0}
\caption{Comparison of CLA method with baseline methods (SAE and SP) for ZSL, average per-class Top-1 accuracy (\%) of unseen classes is reported based on the same data configurations in the different datasets. V to S means from visual to semantic space, while S to V is from semantic to visual space.}
\label{table2}
\begin{center}
\newcommand{\tabincell}[2]{\begin{tabular}{@{}#1@{}}#2\end{tabular}}
\begin{tabular}{lp{2.3cm}p{1cm}p{1cm}p{1cm}p{1.8cm}}
\hline
\bfseries Method &\bfseries Semantic feature &\bfseries AwA &\bfseries CUB &\bfseries Dogs &\bfseries ImNet-2\\
\hline \hline
SAE(V to S)  & att &$78.5$ & $24.0$ & N/A & N/A\\
             & w2v &$48.5$ & $27.7$ & $25.5$ & $11.4$\\
             & glo &$56.0$ & $27.0$ & $20.1$ & N/A\\
             & hie &$59.9$ & $17.9$ & $23.7$ & N/A\\
\hline
SAE(S to V)  & att &$80.3$ & $20.5$ & N/A & N/A\\
             & w2v &$64.7$ & $22.1$ & $31.2$ & $12.7$\\
             & glo &$72.0$ & $26.9$ & $24.9$ & N/A\\
             & hie &$59.9$ & $25.1$ & $25.5$ & N/A\\
\hline
SP     & att &$84.3$ & $51.8$ & N/A & N/A\\
       & w2v &$77.4$ & $32.5$ & $33.3$ & $13.5$\\
       & glo &$70.5$ & $33.3$ & $33.4$ & N/A\\
       & hie &$62.1$ & $24.3$ & $32.4$ & N/A\\
\hline\hline
CLA  & att &$\textbf{86.3}$ & $\textbf{59.5}$ & N/A & N/A\\
     & w2v &$\textbf{80.2}$ & $\textbf{36.2}$ & $\textbf{40.1}$ & $\textbf{13.6}$\\
     & glo &$\textbf{79.2}$ & $\textbf{37.3}$ & $\textbf{38.5}$ & N/A\\
     & hie &$\textbf{73.5}$ & $\textbf{29.2}$ & $\textbf{35.6}$ & N/A\\
\hline
\end{tabular}
\end{center}
\end{table}

\begin{table}[!ht]
\small
\renewcommand{\arraystretch}{1.0}
\caption{Comparison of CLA method with baseline methods (SAE and SP) for ZSL, average per-class Top-n accuracy (\%) of unseen classes are reported based on the same data configurations in ImNet-2. V to S means from visual to semantic space, while S to V is from semantic to visual space.}
\label{table3}
\begin{center}
\newcommand{\tabincell}[2]{\begin{tabular}{@{}#1@{}}#2\end{tabular}}
\begin{tabular}{lp{2.3cm}p{1.8cm}p{1.8cm}p{1cm}p{1cm}}
\hline
\bfseries Top-n &\bfseries Semantic feature &\bfseries SAE(V to S) &\bfseries SAE(S to V) &\bfseries SP &\bfseries CLA\\
\hline \hline
Top-1  & w2v &$11.4$ & $12.7$ & $13.5$ & $\textbf{13.6}$\\
\hline
Top-2  & w2v &$16.7$ & $18.2$ & $18.0$ & $\textbf{19.6}$\\
\hline
Top-3 & w2v &$20.2$ & $21.6$ & $21.4$ & $\textbf{23.3}$\\
\hline
Top-4  & w2v &$22.9$ & $24.4$ & $23.9$ & $\textbf{26.0}$\\
\hline
Top-5   & w2v &$25.3$ & $26.7$ & $26.2$ & $\textbf{28.3}$\\
\hline
\end{tabular}
\end{center}
\end{table}

\subsection{Comparison with existing methods for multi-semantic fusion }
In this section, multi-semantic fusion is implemented by ClA, SJE\cite{7298911}, LatEm\cite{7780384}, SynC \cite{Changpinyo2016} and SP\cite{Lin2018structure}. The details of experimental results are shown in Tab.\ref{table4}. w indicates that multi-semantic fusion includes att, w2v, glo and hie, while w/o expresses that multi-semantic fusion contains w2v, glo and hie. In the different datasets, the classification performance of CLA is better that of other methods. In AWA, CUB, and Dogs, the performance of CLA respectively increases $2.5\%$,$4.7\%$, and $0.5\%$ at least.

\begin{table}[!ht]
\small
\renewcommand{\arraystretch}{1.0}
\caption{Comparison of CLA method with SJE\cite{7298911}, LatEm\cite{7780384}, SynC \cite{Changpinyo2016} and SP\cite{Lin2018structure} for multi-semantic fusion in ZSL, average per-class Top-1 accuracy (\%) of unseen classes are reported based on the same data configurations in the different datasets. w means that multi-semantic fusion includes att, w2v, glo and hie, while w/o expresses that multi-semantic fusion contains w2v, glo and hie.}
\label{table4}
\begin{center}
\newcommand{\tabincell}[2]{\begin{tabular}{@{}#1@{}}#2\end{tabular}}
\begin{tabular}{lp{1cm}p{1cm}p{1cm}p{1cm}p{1cm}p{1cm}p{1cm}}
\hline
\bfseries Dataset &\bfseries Fusion &\bfseries SJE &\bfseries LatEm &\bfseries SynC &\bfseries SP &\bfseries CLA\\
\hline \hline
AWA  & w   &$73.9$ & $76.1$ &$76.2$ & $85.4$ &$\textbf{88.8}$\\
     & w/o &$60.1$ & $66.2$ &$64.5$ & $81.4$ &$\textbf{83.8}$\\
\hline
CUB  & w   &$51.7$ & $47.4$ &$48.5$ & $54.1$ &$\textbf{60.3}$\\
     & w/o &$29.9$ & $34.9$ &$33.6$ & $35.3$ &$\textbf{40.1}$\\
\hline
Dogs & w   &N/A & N/A &N/A & N/A &N/A\\
     & w/o &$35.1$ & $36.3$ &$37.2$ & $48.1$ &$\textbf{48.6}$\\
\hline
\end{tabular}
\end{center}
\end{table}

\subsection{Comparison with state-of-the-arts}
In this section, we compare CLA and state of the arts methods, which include SJE\cite{7298911}, LatEm\cite{7780384}, SynC \cite{Changpinyo2016},SAE\cite{Kodirov2017}, SP\cite{Lin2018structure}, DMaP \cite{Li2017Paths} and AR-CPR\cite{8016672} in Tab.\ref{table5}. The classification performance of CLA outperforms other state of the arts methods except in CUB. When semantic representation is att, DMaP \cite{Li2017Paths} is better than CLA for ZSL. DMaP can focus on the manifold structure consistency between the semantic representation and the image feature, so it can better distinguish unseen classes in semantic representation att. The details of result analysis are explained in section \ref{analysis}.

\begin{table}[!ht]
\small
\renewcommand{\arraystretch}{1.0}
\caption{Comparison of CLA method with state of the arts methods for ZSL, average per-class Top-1 accuracy (\%) of unseen classes are reported based on the same data configurations in the different datasets. w means that multi-semantic fusion includes att, w2v, glo and hie, while w/o expresses that multi-semantic fusion contains w2v, glo and hie.}
\label{table5}
\begin{center}
\newcommand{\tabincell}[2]{\begin{tabular}{@{}#1@{}}#2\end{tabular}}
\begin{tabular}{lp{2.3cm}p{1cm}p{1cm}p{1cm}}
\hline
\bfseries Method &\bfseries Semantic feature &\bfseries AwA &\bfseries CUB &\bfseries Dogs \\
\hline \hline
SJE\cite{7298911}  & att   &$66.7$ & $50.1$ &N/A \\
     & w2v &$51.2$ & $28.4$ &$19.6$ \\
\hline
LatEm\cite{7780384}  & att   &$71.9$ & $45.5$ &N/A  \\
     & w2v &$61.1$ & $31.8$ &$22.6$  \\
\hline
SynC\cite{Changpinyo2016} & att   &$69.3$ & $47.5$ &N/A \\
     & w2v &$52.9$ & $32.3$ &$27.6$ \\
\hline
SAE(V to S)\cite{Kodirov2017} & att   &$78.5$ &$24.0$ &N/A \\
     & w2v &$48.5$ & $27.7$ &$25.5$ \\
\hline
SAE(S to V)\cite{Kodirov2017} & att   &$80.3$ &$20.5$ &N/A \\
     & w2v &$64.7$ & $22.1$ &$31.2$ \\
\hline
SP\cite{Lin2018structure}   & att   &84.3 & $51.8$ &N/A \\
     & w2v &$77.4$ & $32.5$ &$33.3$ \\
\hline
DMaP\cite{Li2017Paths} & att   &$74.9$ & \textbf{61.8} &N/A \\
     & w2v &$67.9$ & $31.6$ &$38.9$ \\
\hline
AR-CPR\cite{8016672} & att   &N/A & $59.5$ &N/A \\
     & w2v &N/A & N/A &N/A \\
\hline\hline
CLA  & att   &$\textbf{86.3}$ & $59.5$ &N/A \\
     & w2v &$\textbf{80.2}$ & $\textbf{36.2}$ &$\textbf{40.1}$ \\
\hline
\end{tabular}
\end{center}
\end{table}

\subsection{Parameter analysis}
\begin{figure*}[ht]
  \begin{center}
\includegraphics[width=0.8\linewidth]{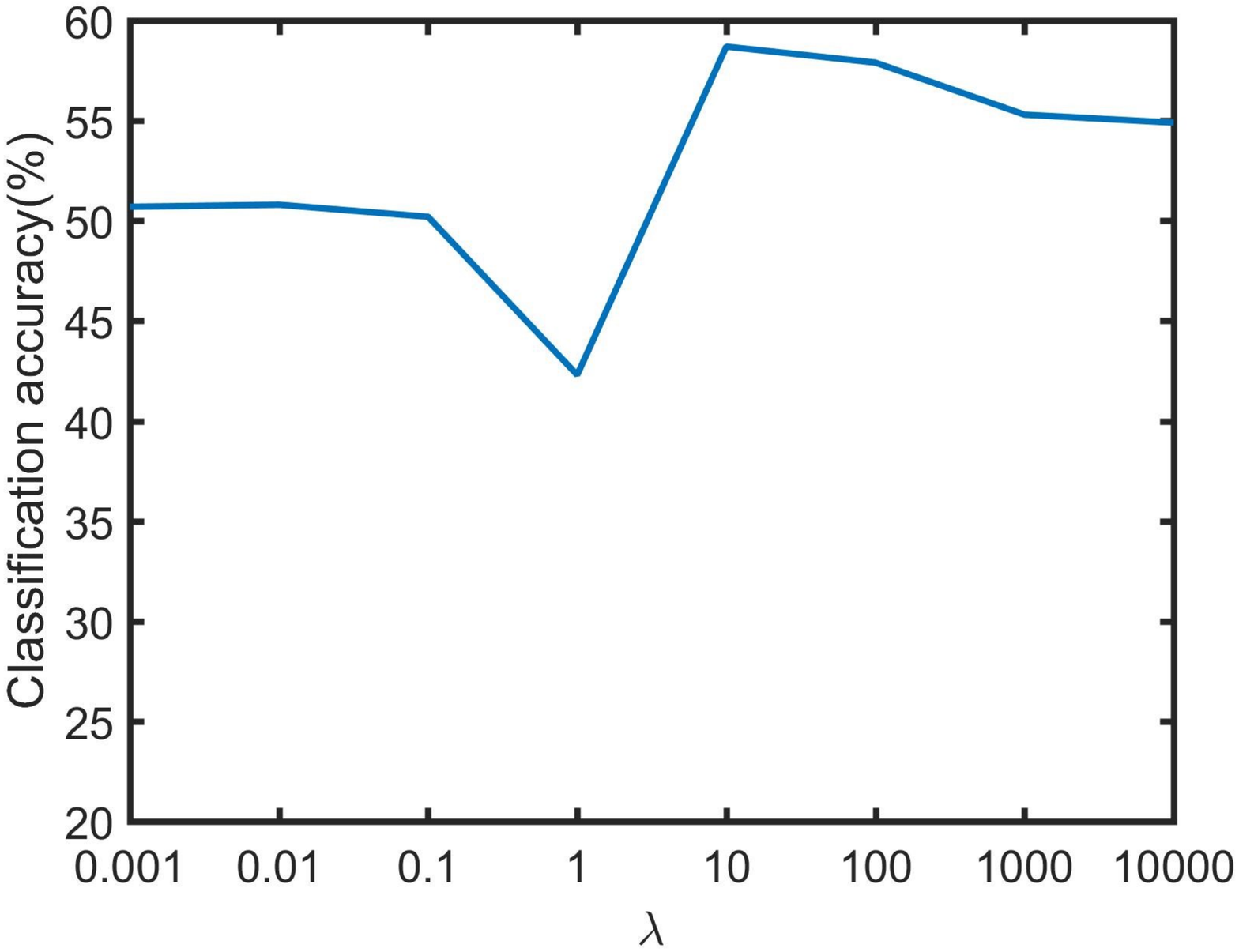}
\end{center}
\vspace{-0.2in}
 \caption{Impact of $\lambda$ on classification performance for zero-shot learning on CUB in att semantic space.}
  \label{fig1}
 \end{figure*}

\begin{figure*}[ht]
  \begin{center}
\includegraphics[width=0.8\linewidth]{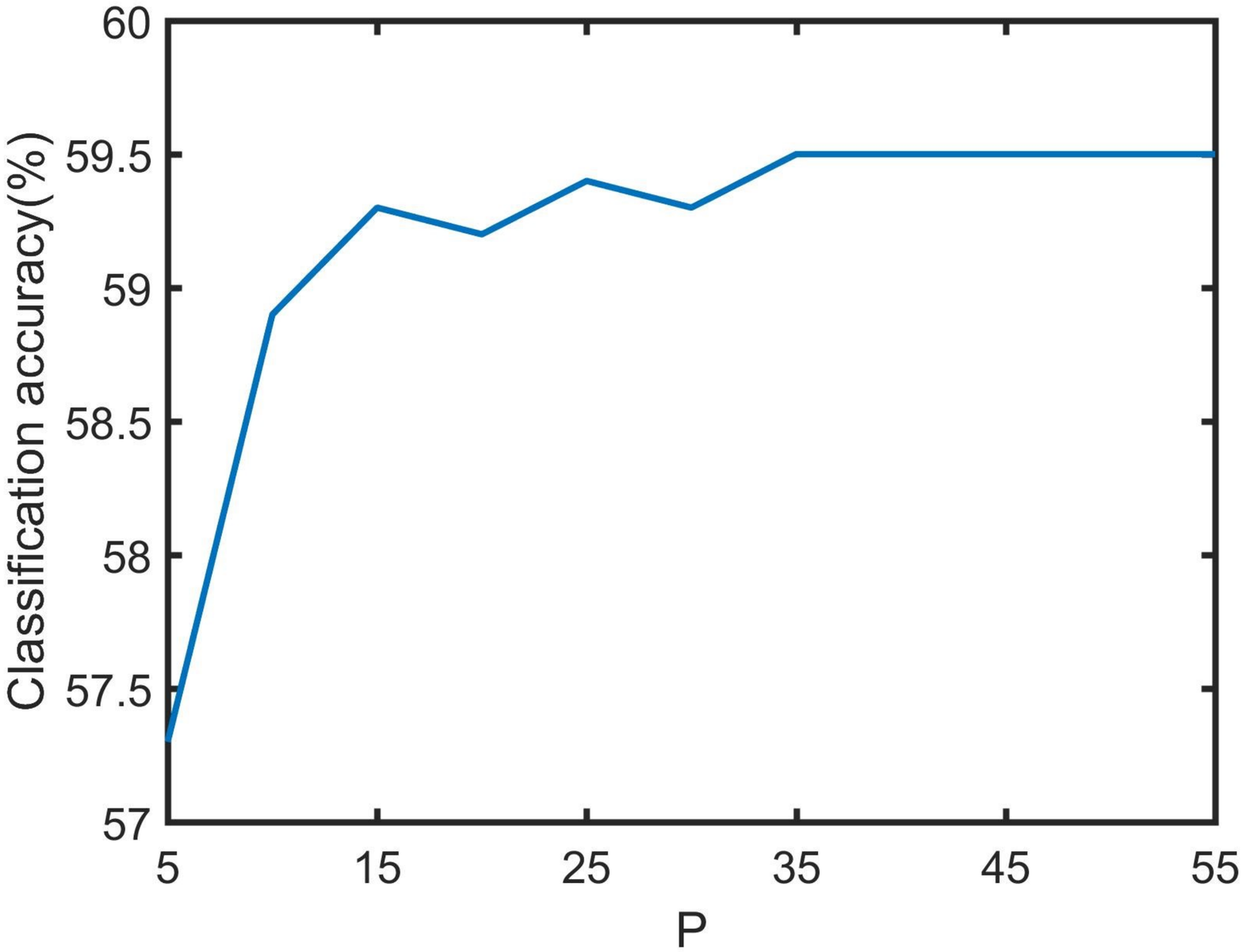}
\end{center}
\vspace{-0.2in}
 \caption{Impact of $P$ on classification performance for zero-shot learning on CUB in att semantic space.}
  \label{fig2}
 \end{figure*}

The important impact on CLA involves two parameter, which are $\lambda$ (the tradeoff parameter and in formula (\ref{objective10}),and $P$ (the total iteration number in Algorithm \ref{algCLA}). As aforementioned, we select $\lambda$ from $0.001,0.01,0.1, 1, 10, 100, 1000, 10000$ by cross-validation on seen classes. In this section, we examine the effect of $\lambda$ on classification performance of CLA in the different selection. Figure \ref{fig1} shows classification performance of CLA with the different $\lambda$ on CUB. We can find that CLA is sensitive to $\lambda$, so we carefully select this parameter by cross-validation in the training set. In above experiments, we empirically set $P$ to 50. However, this parameter indicates the processing of structure evolution. Figure \ref{fig2} shows classification performance of CLA with structure evolution on the different $P$ on CUB. We can observe that classification performance of CLA is improve with $P$ increasing, and then tend to be stable. Therefore, we can set $P$ to 50, because classification performance of CLA is mostly better and stable in this value. Another parameter $\delta$ change is not sensitive to the classification accuracy of CLA.

\subsection{Experimental results analysis}
\label{analysis}
In above experiments, eight approaches are involved to consider the manifold structure from different aspects for building the bridge between visual feature and semantic space. SJE\cite{7298911} can construct the output space of the structure by balancing the different output embedding with the confidence contribution. LatEm\cite{7780384} can capture the structured model for making the overall piecewise linear function, and then can obtain the latent space of the flexible model for fitting the unseen class. SynC\cite{Changpinyo2016} can take into account the manifold structure in semantic space for getting optimal discriminative performance in the model space. SAE\cite{Kodirov2017} can mine the latent manifold structure for classifying unseen classes by taking the encoder-decoder paradigm with the bidirectional constrains. SP\cite{Lin2018structure}can optimize the relationship of the manifold structure in semantic and image space, and improve the positive structure propagation by iteration computation for ZSL. DMaP\cite{Li2017Paths} can build the manifold structure consistency between semantic representation and image feature by using dual visual-semantic mapping paths. AR-CPR\cite{8016672} can learn a deep regression model to matching the manifold structure consistency between images and class prototypes by rectifying class representation. The proposed CLA can jointly consider the structure evolution and the bidirectional constrains between feature and class label space to recognize unseen classes. From aforementioned experiments, we can have the following observations.
\begin{itemize}
\item The performance of CLA outperforms baseline approaches, which are SAE and SP. SAE try to find the mapping relationship with the bidirectional constrains, which are from semantic to image and from image to sematic, while SP attempt to complement the project relevance by structure positive propagation, which implements based on the relationship between unseen and seen classes. The proposed CLA can jointly consider the impact of the the bidirectional constrains, which are from class label to image and from image to class label, and the structure evolution, which involves not only the relevance between unseen and seen classes but also the relationship among unseen classes or seen classes.
\item The performance improvement CLA is different than baseline approaches in the various datasets. The significant advance can be shown in AWA,CUB, and Dogs, while the slight advance can be found in ImNet-2. This situation of the main reason is the diversity of a large-scale dataset to cause the bigger divergence of the intra-class than a small-scale dataset. Therefore, the proposed CLA shows the obvious advantages than other methods in Top-n accuracy experiment.
\item In multi-semantic fusion, the performance of CLA is superior to other methods, which are SJE, LatEm, SynC, and SP. The proposed multi-semantic fusion method can have the better performance than the single-semantic method in term of multi-semantic complement each other. In Dogs dataset, the performance of CLA is slightly better than that of SP, while the performance of CLA significantly outperforms that of other methods. This situation of main reason is that the bidirectional constrains mechanism have the less effect on classification accuracy than structure evolution in multi-semantic fusion methods.
\item In multi-semantic fusion or single-semantic method, structure evolution can have the more improvement the classification accuracy for ZSL than the bidirectional constrains mechanism in supervised semantic space (att). This effect relationship is irregular in unsupervised semantic space (w2v,glo, and hie). However, their joint effect essentially improve the classification accuracy for ZSL.
\item The performance of CLA is superior to that of DMaP except in att semantic space of CUB, because the structure matching of DMaP is a key point for classifying fine-grained category with supervised semantic space (att). CLA integrates the structure evolution and the bidirectional constrains mechanism, while DMaP focuses on the various manifold structure consistency. Therefore, the performance of CLA has approximated to that of DMaP in this situation, even greatly outperforms that of DMaP with unsupervised semantic space (w2v).
\item The performance of CLA and AR-CPR is same in att semantic space of CUB. AR-CPR tempt to train a deep network and rectify class prototype for enhancing classification accuracy of unseen classes, while CLA try to learn the projection function by integrating the structure evolution and the bidirectional constrains mechanism. Both methods obtain the best result based on the different aspects. It shows that class structure distribution and it's constrains are very important to bridge the gap between visual feature and semantic representation.
\item The most computational load involved in CLA is for solving equation (\ref{objective12})and (\ref{objective10}). Specifically,the complexity of equation (\ref{objective12}) is $O((M+1)^{3.5}p)$ ($M$ is semantic space number, and $p$ is the number of bits in the input \cite{Karmarkar1984A}) in the worst case. The complexity of equation (\ref{objective10}) is $O(d^{3})$. Because $M$ and $P$ iteration times are often much less than $d$ feature dimension, the computational complexity of CLA is $O(d^{3})$.
\end{itemize}

\section{Conclusion}
We have proposed class label autoencoder (CLA) method to address multi-semantic fusion in ZSL. CLA can not only adapt multi-semantic structure distribution to a uniform ZSL framework, but also constrain the bidirectional mapping between the feature space and the class label space by the encoder-decoder paradigm. Furthermore, CLA can fuse the relationship of feature classes, the relevance of the semantic classes, and the interaction between feature and semantic classes for improving zero-shot classification. At last, the optimization of the CLA can obtain both unseen class labels and the different classes representation(feature or semantic information) of the relation that can encode the intrinsic structure of classes by iteration evolution way. For evaluating the proposed CLA, we implement the comparison experiments about baseline methods, multi-semantic fusion methods,and state of the art methods on AwA, CUB, Dogs and ImNet-2. Experiment results demonstrate the CLA is effective in ZSL.

\section{Acknowledgements}
The authors would like to thank the anonymous reviewers for their insightful comments that help improve the quality of this paper. Especially, The authors thank to Dr. Yongqin Xian from MPI for Informatics, who provided the data source to us. This work was supported by NSFC (Program No.61771386), Natural Science Basic Research Plan in Shaanxi Province of China (Program No.2017JZ020).

\section*{References}

\bibliography{mybibfile}

\section*{Appendix A}
For optimizing formula (\ref{objective5}), we firstly use trace properties $Tr(X_{s})=Tr(X_{s}^{T})$ and $Tr(A_{s}^{T}W_{s}Y_{s})=Tr(Y_{s}^{T}W_{s}^{T}A_{s})$ to reformulate formula (\ref{objective5}) as following.
 \begin{align}
\label{objective13}
\begin{aligned}
&A_{s}=\arg \min_{A_{s}}\|X_{s}^{T}-Y_{s}^{T}W_{s}^{T}A_{s}\|^{2}_{F}+\lambda\|W_{s}^{T}A_{s}X_{s}-Y_{s}\|^{2}_{F}
 \end{aligned}
\end{align}
Then, we can take a derivative of formula (\ref{objective13}) and set it zero for obtaining the following formula.
 \begin{align}
\label{objective14}
\begin{aligned}
&-Y_{s}(X_{s}^{T}-Y_{s}^{T}W_{s}^{T}A_{s})+\lambda (W_{s}^{T}A_{s}X_{s}-Y_{s})X_{s}^{T}=0\\
&Y_{s}Y_{s}^{T}W_{s}^{T}A_{s}+\lambda W_{s}^{T}A_{s}X_{s}X_{s}^{T}=Y_{s}X_{s}^{T}+\lambda Y_{s}X_{s}^{T}
 \end{aligned}
\end{align}
If $A=Y_{s}Y_{s}^{T}$, $B=\lambda X_{s}X_{s}^{T}$, $C=(1+\lambda)Y_{s}X_{s}^{T}$, and $W=W_{s}^{T}A_{s}$, we can reformulate formula (\ref{objective14}) to the following formula.
\begin{align}
\label{objective15}
\begin{aligned}
&AW+WB=C
 \end{aligned}
\end{align}
Here, formula (\ref{objective15}) is a well-know Sylvester equation.

\end{document}